\documentclass{article}
\usepackage{spconf,amsmath,graphicx}
\usepackage{hyperref}
\usepackage{verbatim}
\usepackage{enumerate}
\usepackage{amsfonts,amsthm, amssymb}
\usepackage{epstopdf}
\usepackage{subfig}
\usepackage{algorithm}
\usepackage{algpseudocode}
\usepackage{algorithmicx}

\title{Structure-Aware Classification using Supervised Dictionary Learning}

\name{Yael Yankelevsky and Michael Elad\thanks{The research leading to these results has received funding from the European Research Council under European Union's Seventh Framework Program, ERC Grant agreement no. 320649, and from the Israel Science Foundation (ISF) grant number 1770/14.}}
\address{Computer Science Department\\Technion - Israel Institute of Technology\\Haifa 32000, Israel}

\begin{document}
%
\maketitle
\begin{abstract}
In this paper, we propose a supervised dictionary learning algorithm that aims to preserve the local geometry in both dimensions of the data.
A graph-based regularization explicitly takes into account the local manifold structure of the data points. A second graph regularization gives similar treatment to the feature domain and helps in learning a more robust dictionary. 
Both graphs can be constructed from the training data or learned and adapted along the dictionary learning process. 
The combination of these two terms promotes the discriminative power of the learned sparse representations and leads to improved classification accuracy.
The proposed method was evaluated on several different datasets, representing both single-label and multi-label classification problems, and demonstrated better performance compared with other dictionary based approaches.
\end{abstract}


\section{Introduction}
Dictionary learning aims to learn a set of atoms such that a given signal can be well approximated by a sparse linear combination of these atoms.
The standard dictionary learning problem is formulated as 
\begin{equation} \label{Eq:KSVD}
\begin{aligned}
&\arg\underset{D,X}{\min}\;\|Y-DX\|_F^2\quad\mbox{ s.t. }\quad \|x_i\|_0\leq T \quad \forall i
\end{aligned}
\end{equation}
where $Y\in\mathbb{R}^{n\times N}$ is the data matrix, $X\in\mathbb{R}^{K\times N}$ contains the sparse representations and $D\in\mathbb{R}^{n\times K}$ is an over-complete dictionary with normalized columns (atoms).

Despite the popularity of standard dictionary learning methods in many domains, their performance in classification tasks is sub-optimal, since an accurate reconstruction is not as important for classification as the discrimination capability of the dictionary. 
This motivates the emergence of supervised dictionary learning techniques, exploiting both the existing label information and the underlying structure of the data.

Some of these methods attempt to learn a separate sub-dictionary for each class (e.g. \cite{Wright2009,Yang2010,Yang2011,Ramirez2010}).
Consequently, the sparse codes over the learned dictionary are used as features on which a classifier is trained. 
More sophisticated approaches (e.g. \cite{Mairal2008,Pham2008,Zhang2010,Jiang2011}) learn a discriminative dictionary by introducing a classification-error term into the objective function, and enforcing some discriminative criteria on the optimized sparse coefficients. By doing so, these methods form a unified learning problem and learn the dictionary and classifier jointly.

Of the latter category, we focus on the Label Consistent K-SVD (LC-KSVD) method \cite{Jiang2011} for joint learning of an over-complete dictionary $D$ and an optimal linear classifier $W$:
\begin{equation} \label{Eq:LC-KSVD}
\begin{aligned}
&\arg\underset{D,W,A,X}{\min}\|Y-DX\|_F^2+\alpha \|Q-AX\|_F^2+\beta\|H-WX\|_F^2 \\
&\quad\mbox{ s.t. }\quad \|x_i\|_0\leq T \quad \forall i,
\end{aligned}
\end{equation}
where $H\in\mathbb{R}^{q\times N}$ is a binary matrix containing the labels of the training data (out of $q$ possible classes), and $Q\in\mathbb{R}^{K\times N}$ associates label information with each dictionary atom, thus forcing signals from the same class to have similar sparse representations.
The minimized objective hence balances between the reconstruction error $\|Y-DX\|_F^2$, the label consistency $\|Q-AX\|_F^2$ and the classification error $\|H-WX\|_F^2$.
These terms can be fused together, leading to a standard formulation:
\begin{equation} \label{Eq:LC-KSVD-fused}
\arg\underset{\tilde{D},X}{\min}\left\|\tilde{Y}-\tilde{D}X\right\|_F^2 \quad \mbox{ s.t. }\quad \|x_i\|_0\leq T \quad \forall i,
\end{equation}
where $\tilde{Y}=\begin{pmatrix}Y\\\sqrt{\alpha} Q\\ \sqrt{\beta} H\end{pmatrix}$ and $\tilde{D}=\begin{pmatrix}D\\\sqrt{\alpha} A \\ \sqrt{\beta} W\end{pmatrix}$.

Equation~\eqref{Eq:LC-KSVD-fused} can be efficiently solved using the K-SVD algorithm \cite{Aharon2006}, which iteratively alternates between a sparse coding step (optimization over $X$) and a dictionary update step (that updates each dictionary atom in $\tilde{D}$ along with its related coefficients from $X$). 
Having completed the training process, the individual components $D$ and $W$ can be recovered from $\tilde{D}$. Consequently, classification of a new signal is simply performed by sparse coding over the dictionary $D$ and applying the learned classifier $W$ on the resulting sparse coefficient vector, choosing the class that yields the highest score.
\newline

In \cite{Yankelevsky2016}, we have suggested an unsupervised dictionary learning algorithm for graph signals. The algorithm takes into account the underlying structure of the data in both the feature and the manifold domains using graph smoothness constraints. Furthermore, the underlying structure, encapsulated by a graph Laplacian matrix, can be learned within the dictionary learning process to promote the desired smoothness. 

In this paper, we propose an extension of our dual graph regularized dictionary learning algorithm to a supervised setting by applying the same ideas to the LC-KSVD approach \cite{Jiang2011}. 
The novelty of the LC-KSVD algorithm lies in the requirement that objects from the same class have similar sparse codes over some dictionary. 
While this method was shown to yield satisfactory classification results, we argue that optimizing a separate sub-dictionary for each class or directly relating the dictionary atoms with specific classes is overly restrictive and highly sensitive to the initialization of the algorithm.
We therefore replace the label consistency constraint with a graph-based smoothness regularization that leverages the label information and promotes the discriminative nature of the sparse codes.
Additionally, we propose to simultaneously take into account the underlying structure of the training data in the feature domain such that the feature dependencies are preserved in the learned dictionary atoms.

In the sequel, we describe the proposed algorithm and demonstrate its efficiency in simulations. 

\section{Graph-Constrained Supervised Dictionary Learning}
\subsection{Introducing the Data Manifold Structure}
Our proposed algorithm is based on the LC-KSVD approach \cite{Jiang2011}.
In order to take into account the local geometrical structure of the data manifold, we shall model the relationships between different data samples using a graph and require smoothness of the sparse codes over the graph topology.

Given a set of training samples $\{y_1,...,y_m\}\in\mathbb{R}^n$, let us construct a weighted graph $\mathcal{M}$ with $m$ vertices, where each node represents a training data point. 
The weight $w_{ij}$ assigned to the edge connecting the $i$-th and $j$-th nodes is designed to be inversely proportional to the distance between them. 
A common choice uses a Gaussian kernel function
\begin{equation}
w_{ij}=\exp\left(-\frac{\|y_i-y_j\|_2^2}{\varepsilon}\right).
\end{equation}
The graph adjacency matrix $W^{\mathcal{M}}$ consists of the edge weights $w_{ij}$. The graph Laplacian $L_{\mathcal{M}}$ is then defined as $L_{\mathcal{M}}=D^{\mathcal{M}}-W^{\mathcal{M}}$, where the degree matrix $D^{\mathcal{M}}$ is a diagonal matrix whose entries are $D^{\mathcal{M}}_{ii}=\sum_j w_{ij}$.

Following ideas from manifold learning and spectral graph theory, the graph Laplacian $L_{\mathcal{M}}$ can be used as a smoothness operator to preserve the local manifold structure. Similarly to the methods proposed in \cite{Zheng2011,Ramamurthy2012}, we incorporate $L_{\mathcal{M}}$ into the objective function as a regularizer of the form $Tr(XL_{\mathcal{M}}X^T)$. 
Denoting the $i$-th column of $X$ by $x_i$, we observe that 
\begin{equation}
Tr(XL_{\mathcal{M}}X^T)=\frac{1}{2}\sum_{i,j}w_{ij}\|x_i-x_j\|_2^2.
\end{equation}
This term therefore encourages similar signals, having a large proximity measure $w_{ij}$, to have similar sparse codes, thus satisfying the commonly known manifold assumption \cite{Belkin2003}. In other words, it promotes smoothness of the obtained sparse representations $X$ along the geodesics of the underlying data manifold, described by the graph Laplacian $L_{\mathcal{M}}$.

Replacing the original label consistency term with the graph regularization, the new dictionary learning problem formulation reads
\begin{equation} \label{Eq:graph-KSVD}
\begin{aligned}
&\arg\underset{D,W,X}{\min}\|Y-DX\|_F^2+\beta\|H-WX\|_F^2+\gamma Tr(XL_{\mathcal{M}}X^T) \\
&\quad\mbox{ s.t. }\quad \|x_i\|_0\leq T \quad \forall i.
\end{aligned}
\end{equation}
Fusing the first two components together similarly to the form presented in Equation~\eqref{Eq:LC-KSVD-fused}, we obtain the graph-regularized supervised dictionary learning problem
\begin{equation} \label{Eq:graph-KSVD-fused}
\begin{aligned}
&\arg\underset{\tilde{D},X}{\min}\left\|\tilde{Y}- \tilde{D}X\right\|_F^2+\gamma Tr(XL_{\mathcal{M}}X^T) \\
&\mbox{ s.t. }\; \|x_i\|_0\leq T \quad \forall i
\end{aligned}
\end{equation}
where $\tilde{Y}=\begin{pmatrix}Y\\ \sqrt{\beta} H\end{pmatrix}$ and $\tilde{D}=\begin{pmatrix}D\\\sqrt{\beta} W\end{pmatrix}$.
\newline

\subsection{Introducing Feature Interdependencies}
In order to take into account the interdependencies in the feature domain as well, we shall construct a second graph to model the relationships between different features, and require smoothness of the dictionary atoms over this new graph.
That is, if two features behave similarly across the training signals, this behavior should be reflected in the structure of the learned atoms.
Explicitly, given the set of training samples $\{y_1,...,y_m\}\in\mathbb{R}^n$, let us construct a weighted graph $\mathcal{G}$ with $n$ vertices, where each node represents a feature (corresponding to a row in the data matrix). 
The weight assigned to the edge connecting the $i$-th and $j$-th nodes is again designed to be inversely proportional to the distance between them. The graph adjacency matrix $W^{\mathcal{G}}$ consists of the edge weights, $D^{\mathcal{G}}$ is the corresponding degree matrix, and the graph Laplacian is defined as $L_{\mathcal{G}}=D^{\mathcal{G}}-W^{\mathcal{G}}$.

In a symmetric view to the manifold graph regularization, the feature graph should be integrated through a regularization term of the form $Tr(D^TL_{\mathcal{G}}D)$ applied to the dictionary matrix $D$. 
Since in the current formulation~\eqref{Eq:graph-KSVD-fused} the dictionary $D$ only occupies a subset of the matrix $\tilde{D}$, we shall zero-pad $L_{\mathcal{G}}$ to the dimension of $\tilde{D}$:
\begin{equation}
\tilde{L}_{\mathcal{G}}=\begin{bmatrix}
L_{\mathcal{G}} & 0\\0 & 0
\end{bmatrix}
\end{equation}
As the extended matrix $\tilde{L}_{\mathcal{G}}$ satisfies $\tilde{D}^T\tilde{L}_{\mathcal{G}}\tilde{D}=D^TL_{\mathcal{G}}D$, the regularization may equivalently be imposed on $\tilde{D}$.

The result is the dual graph constrained dictionary learning problem:
\begin{equation} \label{Eq:graph-KSVD-fused2}
\begin{aligned}
&\arg\underset{\tilde{D},X}{\min}\left\|\tilde{Y}- \tilde{D}X\right\|_F^2+\gamma Tr(XL_{\mathcal{M}}X^T) +\rho Tr(\tilde{D}^T\tilde{L}_{\mathcal{G}}\tilde{D})\\
&\mbox{ s.t. }\; \|x_i\|_0\leq T \quad \forall i
\end{aligned}
\end{equation}
Solving this problem requires significant modifications of the K-SVD algorithm. This can be done using the \textbf{\emph{graph$\bold{^2}$DL}} algorithm we proposed in \cite{Yankelevsky2016} that reflects the added restrictions.
Upon completion of the learning process, $D$ and $W$ can be recovered from $\tilde{D}$, and classification is again performed by sparse coding the test signals using $D$ and applying the classifier $W$ on the resulting coefficients.

\subsection{Learning Dependencies}
The feature dependencies were so far inferred from the patterns detected in the training set. 
For this purpose, an initial weight matrix $W^\mathcal{G}$ for the features graph can be constructed by computing the pairwise distances between rows in the data matrix $Y$, which correspond to the different features:
\begin{equation}
w_{ij}=\exp\left(-\frac{\|Y(i,:)-Y(j,:)\|_2^2}{\varepsilon}\right)
\end{equation}

To better handle partial correlations, these dependencies can be learned and adapted along with the dictionary learning process, as previously suggested in \cite{Yankelevsky2016}.
Having obtained an intermediate dictionary $D$, the graph can be re-estimated based on the dictionary instead of the original input data $Y$, according to the following optimization problem:

\begin{equation} \label{Eq:optimizeL}
\arg\underset{L_{\mathcal{G}}\in\Omega_L^n}{\min}\;\rho Tr(D^TL_{\mathcal{G}}D)+\mu\|L_{\mathcal{G}}\|_F^2,
\end{equation}
where 
\begin{equation} 
\begin{aligned}
\Omega_L^N = \lbrace
L\in\mathbb{R}^{N\times N}\;\vert&\;L=L^T,L_{ij}\leq 0\; (i\neq j),\\
&\;L\underline{1}=\underline{0},\; Tr(L)=N
\rbrace
\end{aligned}
\end{equation}
defines the set of trace-normalized valid graph Laplacian matrices of size $N\times N$. 
The term $\|L_{\mathcal{G}}\|_F^2$ was added to the objective function to control the sparsity of the resulting Laplacian matrix. 
In a similar manner, the manifold graph Laplacian $L_{\mathcal{M}}$ may also be adapted by solving 
\begin{equation} \label{Eq:optimizeLc}
\arg\underset{L_{\mathcal{M}}\in\Omega_L^m}{\min}\;\beta Tr(XL_{\mathcal{M}}X^T)+\eta\|L_{\mathcal{M}}\|_F^2.
\end{equation}

The complete algorithm, as summarized in Algorithm~\ref{Alg:supGraphRegDL}, is assembled by alternating between the graph constrained supervised dictionary learning problem posed in Equation~\eqref{Eq:graph-KSVD-fused2}, and adapting the Laplacian matrices $L_{\mathcal{G}}$ and $L_{\mathcal{M}}$.

\begin{algorithm}[htbp]
	\caption{Graph Constrained Supervised Dictionary Learning}
	\label{Alg:supGraphRegDL}
	\begin{algorithmic} [1]
		\Statex - Initialize $\tilde{D}^{(0)}$ 
		\Statex - Iterate for $k=1,2,...$
		\Statex 
		\begin{itemize}
			\item Run a few iterations of \textbf{graph$\mathbf{^2}$DL} (see \cite{Yankelevsky2016}, Algorithm 3), initialized with $\tilde{D}^{(k-1)}$, to solve Equation~\eqref{Eq:graph-KSVD-fused2} and obtain $\tilde{D}^{(k)},X^{(k)}$. 
			\item Update the graph Laplacians (Equations~\eqref{Eq:optimizeL},\eqref{Eq:optimizeLc}).
		\end{itemize}
		\Statex - Return $\tilde{D}^{(k)},X^{(k)}$
	\end{algorithmic}
\end{algorithm}

\section{Experimental Results}
\subsection{Single-Label Classification}
First, we evaluate the performance of the proposed algorithm for single-label image classification on the AR Face database \cite{AR1998} and on the Extended YaleB database \cite{YALEB2001}.

The AR Face database consists of over 4000 images: 26 images per person for 126 different people. The images feature frontal view faces with different facial expressions, illumination conditions, and occlusions (sun-glasses and scarf).
Following the standard evaluation procedure, we use a subset of the database consisting of 2600 images from 50 male subjects and 50 female subjects. 
Randomly selected 20 images per person constitute the training set, and the rest are used for testing.
Each image is represented by a 540-dimensional feature vector using the procedure described in \cite{Jiang2011}.

The Extended YaleB database contains 2414 frontal-face images of 38 individuals, captured under various illumination conditions. We randomly selected half of the images (about 32 images per person) as the training set and the rest are used for testing.
Each image is represented by a 504-dimensional feature vector.

We compare the proposed algorithm to the two variants proposed in \cite{Jiang2011}: LC-KSVD1, which refers to \eqref{Eq:LC-KSVD} for $\beta=0$, and LC-KSVD2, which refers to \eqref{Eq:LC-KSVD} for $\alpha,\beta\neq 0$.
Two variants of our method are evaluated: \textbf{SupGraphDL}, standing for Algorithm~\ref{Alg:supGraphRegDL} without learning the graph Laplacians $L_\mathcal{G}$ and $L_\mathcal{M}$, and \textbf{SupGraphDL-L}, in which the two Laplacians are adapted throughout the process. 
The obtained results are summarized in Table~\ref{tab:resFaces}.
Note that we used a random partition of the data into training and testing sets, therefore the results obtained for LC-KSVD are different from the best case results presented in \cite{Jiang2011}. 
Nevertheless, our algorithm is able to achieve a higher classification accuracy for both evaluated datasets. 

\begin{table}[htb]
	\centering
	\scalebox{1}{
		\begin{tabular}{|c||c|c|c|c|c|} 
			\hline 
			Method  &  AR & YaleB \\
			\hline \hline
			LC-KSVD1 & 84.17 & 93.12\\
			\hline 
			LC-KSVD2 & 85.00 & 93.29\\
			\hline 
			SupGraphDL & 84.93 & 92.89\\
			\hline 
			SupGraphDL-L & \textbf{85.33} & \textbf{93.44}\\
			\hline 
		\end{tabular} 
	}
	\caption[]{Classification accuracy ($\%$) for the AR Face dataset and the Extended YaleB dataset.} \label{tab:resFaces}
\end{table}

\subsection{Multi-Label Classification}
Next, we evaluate the proposed algorithm for the more challenging task of multi-label classification. In this setting, each instance may be associated with multiple classes simultaneously. Thus, exploiting the interdependency between labels can significantly affect the success of the classification algorithm.

As an initial step, we extended our method to support multi-label classification, by altering the binary label matrix $H$ to allow multiple non-zeros per column.
The classification procedure was also extended to support multiple labels: instead of choosing the class yielding the maximal score, the relevant labels were selected as those reaching a result above a threshold, i.e. $\Omega_i=\left\lbrace\ell \;:\; [Wx_i](\ell)\geq 0.5\right\rbrace$, where $x_i$ is the sparse representation of a test sample $y_i$ over the dictionary $D$, and $W$ is the learned classifier. 
\newline

The algorithm was evaluated for two datasets: natural scene images and yeast gene functionality. 

The natural scene dataset consists of 2000 natural scene images, each belonging to one or more out of 5 possible semantic classes: \emph{desert}, \emph{mountains}, \emph{sea}, \emph{sunset} and \emph{trees}. 
Half of the images were used for training and the rest constitute the test set. 
Each image is represented by a 294-dimensional feature vector using the procedure described in \cite{Boutell2004}. The extracted features are spatial color moments in the LUV space, which are commonly used in the scene classification literature.

The yeast dataset \cite{Elisseeff2001} is formed by micro-array expression data and phylogenetic profiles, and includes 2417 genes, 1500 of which are used for training and the rest constitute the test set. 
Each gene is represented by a 103-dimensional feature vector, and associated with a set of functional groups out of 14 possible classes (such as \emph{metabolism}, \emph{transcription} and \emph{protein synthesis}). 

Similarly to the single-label classification experiment, we compare two variants of our method (with and without adapting the graphs) to LC-KSVD1 and LC-KSVD2 proposed in \cite{Jiang2011}.

To assess the accuracy of the algorithms in the multi-label experiments, we use the average precision measure as defined in~\cite{Boutell2004}. 
The obtained classification results are summarized in Table~\ref{tab:resML}, indicating that our algorithm clearly outperforms the other methods. 

\begin{table}[htb]
	\centering
	\scalebox{1}{
		\begin{tabular}{|c||c|c|c|c|c|} 
			\hline 
			  &  Scene & Yeast \\
			\hline \hline
			LC-KSVD1 & 81.48 & 61.17\\
			\hline 
			LC-KSVD2 & 82.57 & 61.21\\
			\hline 
			SupGraphDL & 82.78 & 64.69\\
			\hline 
			SupGraphDL-L & \textbf{83.80} & \textbf{67.91}\\
			\hline 
		\end{tabular} 
	}
	\caption[]{Classification accuracy ($\%$) for the multi-label Scene and Yeast datasets.} \label{tab:resML}
\end{table}

In the multi-modal scenario, the potential overlap between different combinations of classes implies that a more complex underlying structure could be learned and exploited. 
As expected, the impact of the structural constraints in the multi-modal scenario is very significant, and more pronounced compared with the single-label examples. 
For the yeast dataset, which is known to be difficult, the achieved improvement in classification accuracy is almost $7\%$.
Moreover, though increasing the computational complexity, it can be observed that adapting the Laplacian matrices further improves the classification results for both datasets.

\section{Conclusions}
In this paper, we suggested an extension of our previously proposed dual graph regularized dictionary learning algorithm to a supervised setting. 
In the new algorithm, the dictionary atoms are encouraged to preserve the feature similarities as detected in the training data and encapsulated by the graph Laplacian $L_\mathcal{G}$, thus leading to a better representative and more robust dictionary.
By adhering to the intrinsic geometrical structure of the data manifold, as captured by the graph Laplacian $L_\mathcal{M}$, the resulting sparse codes are more discriminative and can significantly enhance classification performance. 

Experiments performed on different datasets demonstrate that the proposed algorithm yields very good classification results, outperforming other supervised dictionary learning algorithms for both single-label and multi-label classification tasks.



\bibliographystyle{IEEEbib}
\bibliography{learning_refs}
\end{document}